%% file: main.tex
\documentclass{article}
\usepackage[preprint,nonatbib]{arxiv}
\usepackage{graphicx} % Required for inserting images
\usepackage{hyperref}
\usepackage{amsfonts,amssymb}
\usepackage{amsmath}
\usepackage{algorithm}
\usepackage{algpseudocode}
\usepackage[dvipsnames]{xcolor}
\usepackage[utf8]{inputenc}
\usepackage[T1]{fontenc}
\usepackage{algpseudocode}
\usepackage{makecell}
\usepackage{url}
\usepackage{tabularx}
\usepackage{caption}

\title{TQCompressor: improving tensor decomposition methods in neural networks via permutations}
\author{
  \textbf{V. Abronin$^1$, A. Naumov$^1$, D. Mazur$^1$, D. Bystrov$^1$, K. Tsarova$^1$,} \\
  \textbf{Ar. Melnikov$^1$, I. Oseledets, S. Dolgov$^1$, R. Brasher$^1$, M. Perelshtein$^{1, \dagger}$} \\[3 pt]
  $^1$\,Terra Quantum AG, Kornhausstrasse 25, 9000 St. Gallen, Switzerland \\[3 pt]
  $^{\dagger}$\texttt{mpe@terraquantum.swiss}
}

\date{August 2023}

\begin{document}

\maketitle

\begin{abstract}
We introduce TQCompressor, a novel method for neural network model compression with improved tensor decompositions.
We explore the challenges posed by the computational and storage demands of pre-trained language models in NLP tasks and propose a permutation-based enhancement to Kronecker decomposition. This enhancement makes it possible to reduce loss in model expressivity which is usually associated with factorization.  We demonstrate this method applied to the GPT-2$_{small}$ \cite{gpt2}.
The result of the compression is TQCompressedGPT-2 model, featuring 81 mln. parameters compared to 124 mln. in the GPT-2$_{small}$. 
We make TQCompressedGPT-2 publicly available.
We further enhance the performance of the TQCompressedGPT-2 through a training strategy involving multi-step knowledge distillation, using only a 3.1\% of the OpenWebText \cite{openwebtext}. % RB TODO: Search and replace
TQCompressedGPT-2 surpasses DistilGPT-2 \cite{distilbert} and KnGPT-2 \cite{kngpt2} in comparative evaluations, marking an advancement in the efficient and effective deployment of models in resource-constrained environments.
\end{abstract}

\input{introduction}

\input{related-works}

\input{methodology}

\input{experiments}

\input{conclusion}

\newpage
\input{references}

\newpage
\input{appendix}

\end{document}

%% file: introduction.tex
\section{Introduction}

Advancements in pre-trained language models have significantly impacted the field of natural language processing (NLP), providing robust solutions for tasks such as language translation, text summarization, and question answering. Despite their effectiveness, the substantial computational and storage demands of these models present a notable challenge. Consequently, research efforts have increasingly concentrated on model compression strategies to alleviate these demands. Techniques such as knowledge distillation \cite{distilling_the_knowledge_in_neural_network}, post-training quantization \cite{int8}\cite{smoothquant}, pruning \cite{int8}, and matrix factorization, specifically tensor decomposition \cite{oseledets_tensorizing_embedding_layers} \cite{novikov_tensorizing_neural_networks}, are at the forefront of these endeavors.

Matrix factorization methods offer promising prospects, as demonstrated by the high compression ratios achieved without significant performance drops in downstream tasks, such as KroneckerBERT's \cite{kroneckerbert} success on the GLUE benchmark \cite{glue}. While these methods theoretically reduce FLOPs, the full realization of these benefits is hindered by current hardware and software limitations, which are not optimized for sparse structures resulting from matrix factorization.

The future, however, looks bright with potential solutions on the horizon. Advances in computing architecture, particularly in quantum computing and adapted GPU designs, are set to bridge this gap. Quantum computing, in particular, with its proficiency in handling complex computations, aligns well with the demands of matrix factorization. Research in this area, such as developments in tensor network computational architectures \cite{qu2021_tt_hardware} \cite{lykov2022tensor}, is already paving the way. These innovations suggest that the theoretical advantages of matrix factorization could soon be fully realized,

Some works focus specifically on factorizing weights in Language Models (LMs) using Kronecker decomposition\cite{kronecker_product_with_applications}. They have been applied to such models as BERT \cite{kroneckerbert} and GPT-2 \cite{kngpt2}, which are reliable model compression benchmarks. Kronecker decomposition can drastically reduce the models number of parameters at the expense of its expressivity. This drawback is then corrected by training the model on a fraction of the original dataset.

This paper introduces a novel permutation-based enhancement to the Kronecker decomposition method, reducing the drop in model expressivity commonly associated with factorization. Our technique, applied to the GPT-2$_{small}$ model—a standard in compression benchmarks—modifies the embeddings, multi-head attention (MHA), and feed forward neural network (FFN) modules. We replace these modules with customized, compressed layers, resulting in a streamlined model with fewer parameters yet comparable performance.

Our method demonstrates its versatility by being applicable to various neural network matrix decomposition techniques. By applying permutations to the weight matrices, we rearrange neuron connections, optimizing the structure for tensor decomposition without altering the network's fundamental functionality. This rearrangement enhances the network's representational power and accuracy.

We further enhance our model's performance using a training strategy based on knowledge distillation \cite{distilling_the_knowledge_in_neural_network}, which effectively mitigates the effects of compression while utilizing only a selected small subset of the dataset (3.1\% of OpenWebText). Our approach involves an epoch-by-epoch decomposition technique during training, which incrementally optimizes the model's structure. In comparative evaluations, this method surpasses its direct competitors and shows superior performance compared to DistillGPT-2 \cite{distilbert}.

\textbf{Our contribution:}

\begin{itemize}
    \item We introduce a new permutation algorithm designed to enhance the efficacy of matrix factorization methods. This algorithm theoretically reduces the performance degradation typically associated with these methods, thereby improving model efficiency without compromising on accuracy.
    \item Our method's effectiveness is showcased through its application to the Kronecker Decomposition method, specifically in the context of compressing the GPT-2$_{small}$ model. This demonstrates not only a reduction in model size but also a maintenance of high performance levels, indicative of the algorithm's potential applicability across various neural network architectures.
    \item We have made the trained weights of TQCompressedGPT-2 available along with the code\footnote{The code and model are publicly available: \\ \url{https://huggingface.co/tq-ag/TQCompressedGPT2}\\ \url{https://github.com/terra-quantum-public/TQCompressedGPT2}}
\end{itemize}

\begin{figure}
    \centering
    \includegraphics[width=0.8\textwidth]{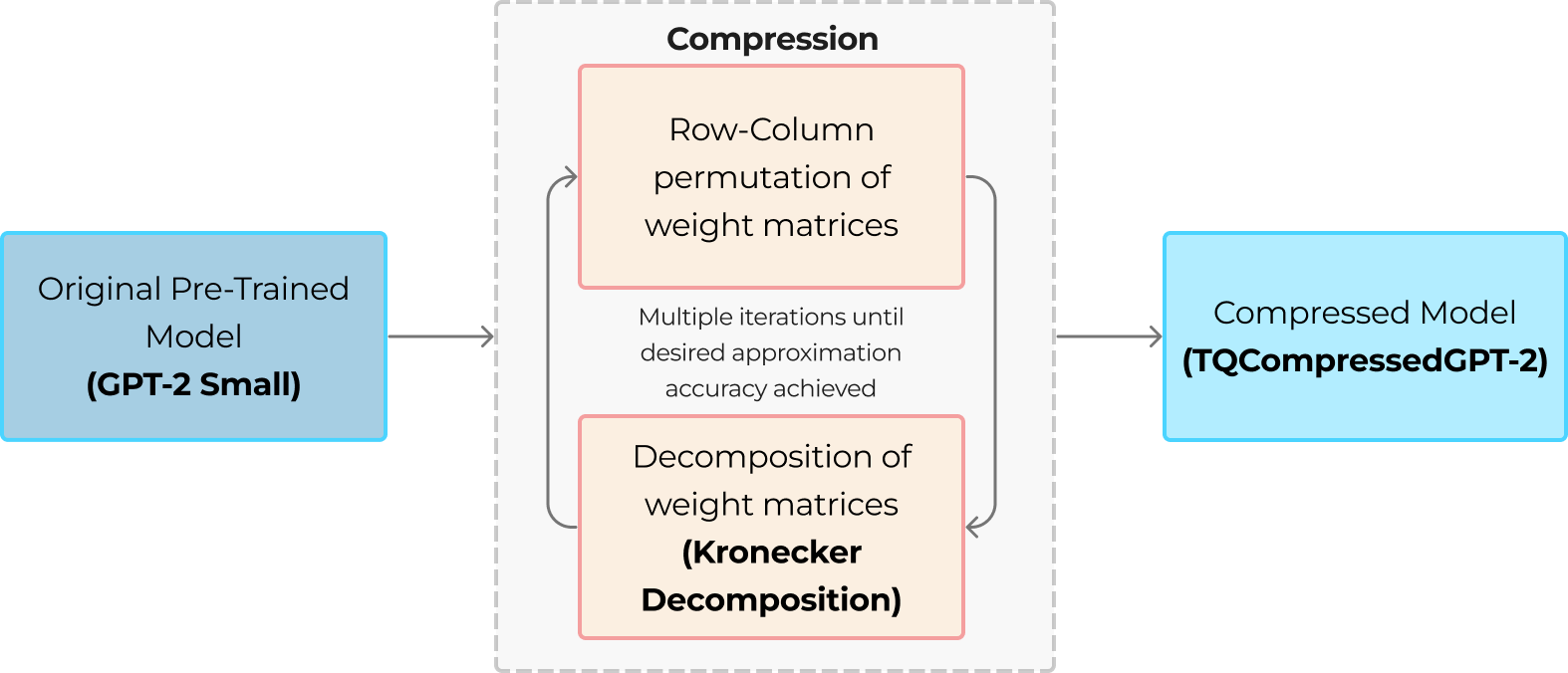}
    \captionsetup{font=small, width=0.8\textwidth, indention=0.05\textwidth} 
    \caption{This figure illustrates the compression pipeline of a pre-trained GPT-2$_{small}$ model using the our Decomposition algorithm. The process begins with the original, uncompressed model on the left. The compression algorithm is applied in the central part of the diagram, consisting of three main steps: First, the row-column permutation of weight matrices is performed to improve the representability of the matrices for decomposition. Second step involves the Kronecker Decomposition of the permuted weight matrices. Algorithm performs multiple iterations until the desired level of approximation accuracy is achieved. The outcome of this process is the compressed model -- TQCompressedGPT-2, shown on the right side of the diagram. This compressed model retains essential performance characteristics while reducing the overall number of parameters, thereby making it more efficient for deployment and use in resource-constrained environments.}
    \label{fig:general_scheme}
\end{figure}

%% file: related-works.tex
\section{Related works}

In this section, we review the related works in the field, with a focus on the Kronecker decomposition algorithm combined with permutations of weight matrices, which is the approach employed in our work. 

Matrix factorization methods have gained prominence as efficient techniques for compressing neural networks. These methods aim to approximate the weight matrices of neural networks with a lower-rank factorization, reducing both the number of parameters and theoretical computational complexity. Matrix factorization approaches for neural network compression include, but not limited to singular value decomposition (SVD) \cite{svd}, Tensor Train decomposition \cite{tensor_train}, Tensor Ring \cite{tensorring} and Kronecker decomposition \cite{kronecker_product_with_applications}.

Numerous studies have focused on the Kronecker decomposition of weights, primarily due to their ability to achieve high compression rates while maintaining, or minimally impacting, model perplexity. The Kronecker decomposition stands out not only for its compression efficacy but also for its significant reduction in FLOPs, which is a crucial factor in enhancing computational efficiency.

A notable milestone in this domain was set by Edalati and Tahaei (2021)\cite{kngpt2}, who pioneered the use of Kronecker decomposition for compressing the GPT-2$_{small}$ model. This model, containing approximately 83 million parameters, is versatile, having been pre-trained under multiple setups to adapt to various downstream tasks. Their work demonstrated not just a feasible approach to compressing a complex model like GPT-2 but also opened avenues for further research in applying such decompositions more broadly across different model architectures. This breakthrough is particularly significant given the increasing demand for efficient, high-performing models in a wide range of applications, from natural language processing to more computationally intensive tasks.

While factorized models generally exhibit an increase in perplexity, this can be effectively countered by fine-tuning the factorized model using a subset of the original dataset. A prevalent strategy for this fine-tuning process is Knowledge Distillation, as outlined in \cite{distilling_the_knowledge_in_neural_network}.

In our research, we also analyze DistilGPT-2 \cite{distilbert}, a prominent compressed variant of the GPT-2$_{small}$ model. DistilGPT-2, with its 82 million parameters, presents a more compact alternative to the 124 million parameters of the original GPT-2$_{small}$, illustrating the potential of model compression techniques in maintaining performance while reducing size. Notably, DistilGPT-2 was trained via Knowledge Distillation using the entire OpenWebText dataset \cite{openwebtext}, which exemplifies the effectiveness of this approach in creating efficient yet powerful models. In contrast, our method demonstrates its efficiency by requiring only small portion of the dataset to recover expressivity drop, significantly reducing the resources needed while still achieving comparable performance.

%% file: methodology.tex
\section{Methodology}

In this section, we expound on our contributions to the field of neural network compression, detailing both the decomposition method and the corresponding training procedure. Let $W \in \mathbb{R}^{m \times n}$ denote a weight matrix within the original neural network architecture. $P \in \mathbb{R}^{m \times m}$ and $C \in \mathbb{R}^{n \times n}$ represent the learned permutation matrices for rows and columns, respectively, while $A \in \mathbb{R}^{m_1 \times n_1}$ and $B \in \mathbb{R}^{m_2 \times n_2}$ describe the Kronecker decomposition matrices for the weight matrix $W$.

\subsection{TQCompressed decomposition}
\label{tqcompressed_decomposition}

Building upon the premise that Kronecker products can effectively compress weights across diverse neural network architectures, we present a novel decomposition approach. This method focuses on identifying an optimal permutation of the weight matrix, followed by its Kronecker decomposition (Figure \ref{fig:general_scheme}). This problem can be mathematically formulated as:

\begin{equation}
\min_{A_i, B_i, P, C} | P W C - \sum_{i=1}^{r} A_i \otimes B_i |_2^2
\end{equation}

It is crucial to note that the permutation matrices $P$ and $C$ are binary matrices, each with a single non-zero element per row, and thus they can be succinctly expressed as permutation vectors of dimensions $n$ and $m$, respectively. The introduction of permutations into our method incurs an additional parameter cost of $n + m$ for each decomposed matrix, resulting in following total parameter count of decomposed matrix: 
\begin{equation}
m_1 n_1 + m_2 n_2 + n + m
\end{equation}
The underlying intuition of our method is akin to shuffling neurons: permuting neurons does not alter the network's structural integrity but renders it more amenable to factorization, enabling a more efficient approximation of each layer.

\subsection{Finding optimal decompositions}
\label{finding_optimal_decompositions}

Determining optimal decompositions is challenging due to the discrete nature of permutation matrices. Our iterative algorithm (Algorithm \ref{alg:iterative_compression}) alternates between optimizing for $P$, $C$ (permutation matrices), and $A$, $B$ (Kronecker decomposition matrices), refining weight approximations.

Optimal permutations are determined in isolation, while the Kronecker matrices $A$ and $B$ are optimized jointly. Each step concludes by fixing the newly obtained optima, and the algorithm proceeds iteratively, optimizing one argument at a time to converge to a solution close to the global optimum.

Initial values for $A$ and $B$ are randomly generated, while $P$ and $C$ begin as identity matrices. The subsequent sections detail each stage of the algorithm.

\subsubsection{Optimal permutation matrices}
\label{optimal_permutation_matrices}

In this section we state the problem of finding optimal permutation matrices and explain our solution of it. Explanations are only given for the row-permutation matrix $P$, but the same reasoning can be applied to the column-permutation matrix $C$. The problem of finding an optimal permutation matrix $P$ can be formulated as follows:

\begin{equation}
\min _{ {P}}
\| {P} {W} {C}- {A} \otimes  {B}\|^2_2
\end{equation}

For simplicity we denote the $WC$ product and the Kronecker product $A \otimes B$ as $W^{(1)}$ and $W^{(2)}$ respectively:

\begin{equation}
\min _{ {P}}
\| {P} {W^{(1)}}- {W^{(2)}}\|^2_2
\end{equation}

In this setting the permutation matrix $P$ can be thought of as a bijective mapping that pairs rows from $W^{(1)}$ to rows from $W^{(2)}$. Thus the problem is reduced to finding a one-to-one correspondence between the rows of $W^{(1)}$ and $W^{(2)}$ that minimizes the total mean squared error between paired rows. This problem statement is equivalent to the assignment problem \cite{kuhn1955hungarian} with the following cost matrix \ref{theorem}:

\begin{equation}
\label{assignment}
    D_{ij} = \displaystyle\sum_{k}|W^{(1)}_{ik} -W^{(2)}_{jk}|,\ D\in \mathbb{R}^{n\times n}
\end{equation}

We solve the assignment problem using the Hungarian algorithm \cite{hungarian, kuhn1955hungarian}.

\begin{algorithm}
\begin{algorithmic}
\caption{Hungarian algorithm for solving \ref{assignment}}\label{alg:cap}

\State // initialize\ matrices
\State $A := rand\_matrix(m_1\times n_1)$
\State $B := rand\_matrix(m_2\times n_2)$
\State $P := id\_matrix(m)$
\State $C := id\_matrix(n)$

\For{k iterations}
\State $A,B:=kron\_decomp(PWC)$ //kronecker decomposition using SVD
\State $Dp:= find\_D(W\cdot C, A\otimes B)$
\State $P:= hung\_alg(Dp)$ //hungarian algorithm 
\State $Dc:= find\_D(W ^ T\cdot P ^ T, (A\otimes B) ^ T) ^ T$
\State $C:= hung\_alg(Dc)$ //hungarian algorithm 

\EndFor

\end{algorithmic}
\end{algorithm}

\subsubsection{Optimal Kronecker decomposition}

When the optimal values for $P$ and $C$ are found, we can find the optimal values for $A$ and $B$ by solving the following problem:

\begin{equation}
    \min_{ {A_i}, {B_i} }
    \|W_{perm} - \sum_{i=1}^{r}A_i\otimes B_i\|^2_2
\end{equation}

Here $W_{perm}$ is the product $WPC$ with fixed optimal values for $P$ and $C$. This problem can be solved using rank-1 SVD decomposition\cite{svd}. We find that our method outperforms vanilla Kronecker decomposition, while adding an insignificant overhead to the parameter count and inference costs.

\subsection{TQCompressed embedding}

Embedding layers are fundamental in natural language processing (NLP) applications, serving as sizable, trainable lookup tables that map discrete tokens to continuous vector spaces. These layers are mathematically characterized by an embedding matrix $W_{emb} \in \mathbb{R}^{v \times d}$, where $v$ is the vocabulary size and $d$ is the embedding dimensionality.

In the architecture of GPT models, two embedding layers are utilized: one for token representation and another for positional encoding within the sequence. Our compression technique targets the weight matrices of these embedding layers along the embedding axis. We apply Kronecker product factorization in the form of $A^{v \times d / f} \otimes B^{f \times d}$, which allows us to represent the original embedding matrix $W_{emb}$ as the product of two smaller matrices $A$ and $B$ with a reduction factor $f$.

\subsection{TQCompressed transformer}

The transformer architecture, pivotal to modern NLP models, is composed of two main elements: multi-head attention (MHA) and feed-forward network (FFN) layers. Our compression framework encompasses both these components.

For the MHA layers, the attention mechanism is realized by first projecting the input through three sets of weights—$W^Q$, $W^K$, and $W^V$—to create the query, key, and value matrices, respectively. The attention output $O$ is computed by the equation:

\begin{equation}
\begin{aligned}
O &= \text{softmax}\left(\frac{QK^T}{\sqrt{d_k}}\right)V,
\end{aligned}
\end{equation}

where $d_k$ is the scaling factor, typically the dimensionality of the Key vectors.

Each head in an MHA layer has its individual set of weight matrices $W^Q_l$, $W^K_l$, and $W^V_l$, which are conventionally concatenated across heads:

\begin{equation}
\begin{aligned}
W^{'Q} &= \text{Concat}(W_1^Q, \ldots, W_L^Q), \\
W^{'K} &= \text{Concat}(W_1^K, \ldots, W_L^K), \\
W^{'V} &= \text{Concat}(W_1^V, \ldots, W_L^V),
\end{aligned}
\end{equation}

where $L$ signifies the total number of attention heads.

In our approach, we perform Kronecker-based decomposition on these concatenated weights as a whole, rather than individually per head:
\begin{equation}
\begin{aligned}
W^{'Q} &\approx P^Q(A^Q \otimes B^Q)C^Q, \\
W^{'K} &\approx P^K(A^K \otimes B^K)C^K, \\
W^{'V} &\approx P^V(A^V \otimes B^V)C^V,
\end{aligned}
\end{equation}

Here, $P^Q$, $P^K$, and $P^V$ are the row permutation matrices, and $C^Q$, $C^K$, and $C^V$ are the corresponding column permutation matrices for the query, key, and value weights, respectively. This factorization allows the model to maintain a high degree of representational fidelity with a reduced number of parameters.

The output of the MHA layers is then projected using another linear transformation $W^O$, which, along with the subsequent FFN weights $W_{1}$ and $W_{2}$, is also subject to our compression algorithm:

\begin{equation}
\begin{aligned}
W^O &\approx P^O(A^O \otimes B^O)C^O, \\
W_{1} &\approx P_1(A_1 \otimes B_1)C_1, \\
W_{2} &\approx P_2(A_2 \otimes B_2)C_2,
\end{aligned}
\end{equation}

\begin{figure}
    \centering
    \includegraphics[width=0.8\textwidth]{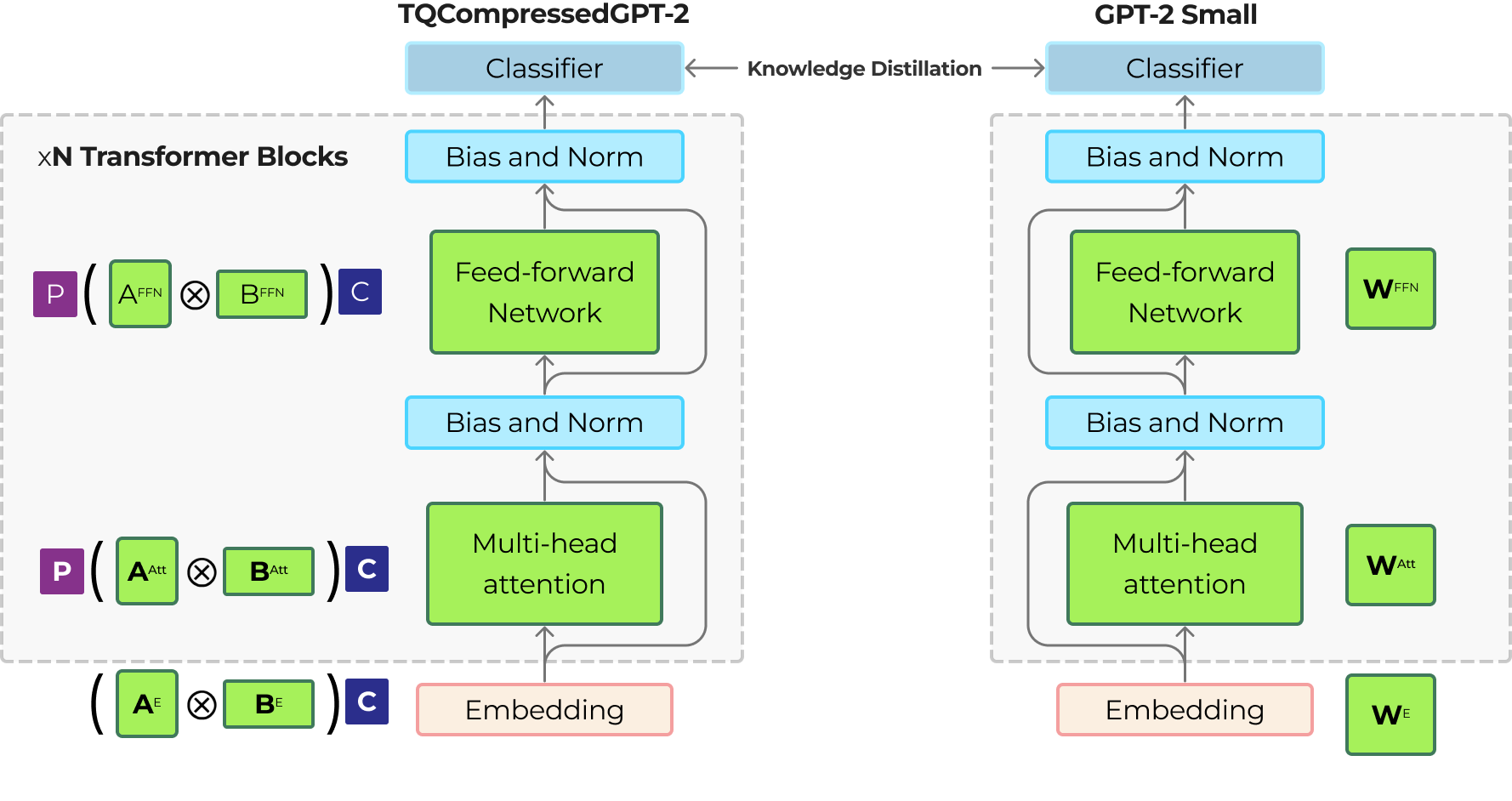}
    \captionsetup{font=small, width=0.8\textwidth, indention=0.05\textwidth} 
    \caption{Only specific layers undergo a compression process — embedding Layer (E), feed-forward network (FFN), and multi-head attention layer (MHA). This process involves applying row-permutation (P) and column-permutation (C) matrices to the original weight matrices, followed by Kronecker decomposition, represented by matrices A and B. The classifier outputs from both the original and the compressed models are then used in a knowledge distillation framework, where the original GPT-2$_{small}$ model serves as the teacher, and TQCompressedGPT-2 acts as the student. The distillation process is focused on aligning the classifier outputs, thereby preserving the performance of the compressed model relative to its original counterpart.}
    \label{fig:tq-compressed-gpt2}
\end{figure}

Figure. \ref{fig:tq-compressed-gpt2} illustrates the GPT-2$_{small}$ weight matrix decomposition scheme, followed by the knowledge distillation, disclosed in detail in the section below.

\subsection{Knowledge distillation}
\label{knowledge distillation}
This section delineates the application of knowledge distillation (KD) \cite{distilling_the_knowledge_in_neural_network} in the training process of compressed model. Within the scope of our discussion, the symbols $S$ (student model) and $T$ (teacher model) are used for representational clarity.

Knowledge distillation was employed by minimizing the divergence between the probability distributions of the student's predictions and the soft targets provided by the teacher's predictions. Concretely, we minimized the cross-entropy loss between the outputs (logits) of the student and teacher models.

The composite loss function amalgamates the standard cross-entropy loss and an additional term representing the cross-entropy of the logits:
\begin{equation}
\mathcal{L}{_\text{total}}(x, y) = \lambda \mathcal{L}{_\text{Cross-Entropy}}(S(x), y) + (1 - \lambda) \mathcal{L}_{\text{Logits}}(S(x), T(x))
\end{equation}
In the loss function $\mathcal{L}_{\text{total}}$, the variable $x$ denotes the input data 
that the neural network processes, while $y$ stands for the corresponding correct labels 
or targets that the network is intended to predict. The loss function evaluates the 
performance of the student model $S$ by comparing its predicted output $S(x)$ against the 
true labels $y$ and the teacher model's predictions $T(x)$, enabling the student to learn 
and approximate the teacher's behavior more effectively. $\lambda$ denotes a balancing coefficient

\begin{figure}[h!]
    \centering
    \includegraphics[width=0.8\textwidth]{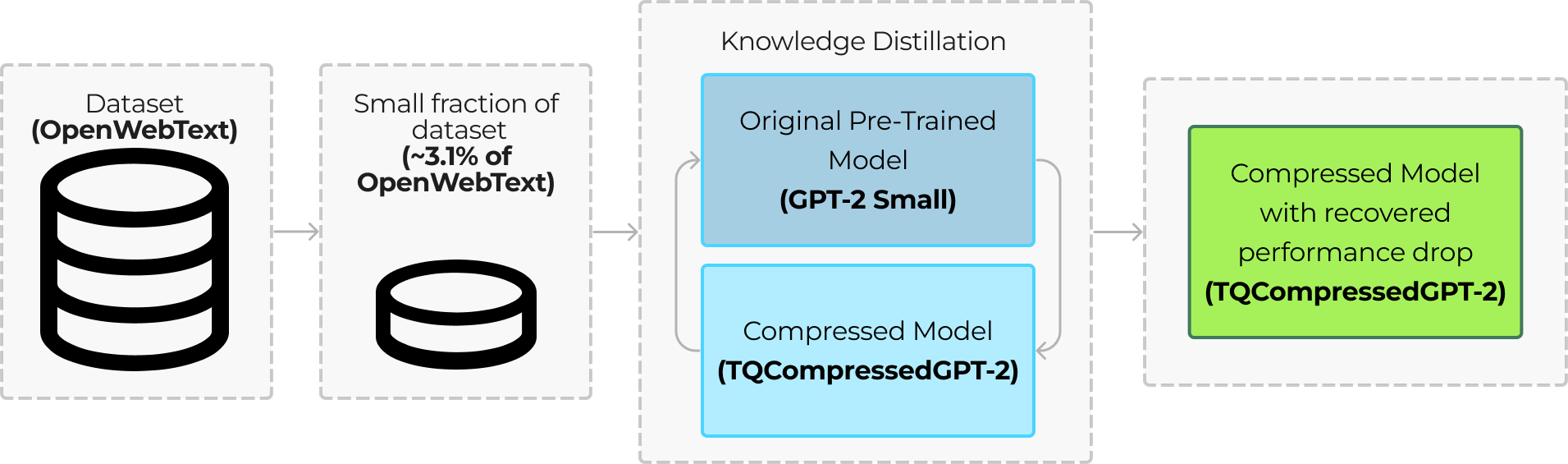}
    \captionsetup{font=small, width=0.8\textwidth, indention=0.05\textwidth} 
    \caption{Knowledge distillation process for compressing the GPT-2$_{small}$ model into the TQCompressedGPT-2 variant. A distilled subset of the OpenWebText corpus serves as the training data for the TQCompressedGPT-2, the student model, which learns under the guidance of the full-sized GPT-2$_{small}$, the teacher model. This process is designed to preserve high performance in the compressed model by closely matching the teacher's accuracy.}
    \label{fig:kd}
\end{figure}

In a departure from the typical implementation of knowledge distillation, our approach harnesses only a fractional subset of the dataset—approximately 3.1\% of the original OpenWebText dataset. This strategy significantly enhances training efficiency.

Figure \ref{fig:kd} visualizes the tailored knowledge distillation framework employed in our study.

\subsection{Iterative compression with knowledge distillation}

The initial application of model compression often results in a non-trivial performance deficit. To mitigate these effects, previous research has leveraged knowledge distillation (KD) \cite{kroneckerbert} \cite{kngpt2}. Our methodology advances this technique by sequentially compressing different layers at various stages throughout the training process. This iterative approach allows the model to dynamically adjust and recalibrate to the compression of individual layers, which we have found to be instrumental in achieving more rapid convergence and improved overall model performance. This iterative process, coupled with knowledge distillation, enables the rest of the model to adapt seamlessly to the modifications in the compressed layers.

Algorithm \ref{alg:iterative_compression} illustrates process for our iterative decomposition with permutations:

\begin{algorithm}
\caption{Iterative decomposition with permutations and knowledge distillation}\label{alg:iterative_compression}
\begin{algorithmic}
\State Let $\theta$ denote the model weights
\State Let $L$ denote the set of all layers in the model
\State Let $L_{\text{C}}$ be the subset of $L$ containing only layers which we want to compress
\For{$i \gets 1$ to $\text{num\_iter}$}
    \State Identify the $i$-th subset of layers $L_i \subseteq L_{\text{C}}$ for compression
    \State Find optimal row and column permutations $P$, $C$ for layers in $L_i$
    \State Apply Kronecker decomposition to obtain $A_i$ and $B_i$ for each layer in $L_i$
    \State Update $\theta$ to reflect compressed layers using $P$, $C$, $A_i$, and $B_i$
    \State Train the compressed model using knowledge distillation
    \State Update $\theta$ with the learned weights from distillation
\EndFor
\State \Return $\theta$
\end{algorithmic}
\end{algorithm}

The $\text{num\_iter}$ parameter specifies the total number of iterations in the compression and training cycle, with each iteration targeting different model layers for compression. This granular approach not only refines the compression process but also aligns the student model more closely with the teacher's performance.

%% file: experiments.tex
\section{Experiments}

In our study, we focus on training a compressed version of the GPT-2$_{small}$ model, which we refer to as TQCompressedGPT-2. This model undergoes training on approximately 3.1\% of the OpenWebText dataset, a significantly smaller subset compared to the full dataset typically used. We then evaluate our model's performance on the Causal Language Modeling task, assessing its perplexity on benchmark datasets such as Wikitext-2, Wikitext-103\cite{wikitext}, and Lambada\cite{lambada}.

\subsection{Experimental setup}

\begin{itemize}
\item \textbf{Pre-training dataset}: We utilize 250K texts from the OpenWebText dataset.
\item \textbf{Downstream task}: The primary benchmarking task is causal language modeling (CLM), with evaluations conducted on the Wikitext-2, Wikitext-103, and Lambada datasets.
\item \textbf{Decomposition shapes}: Our decomposition strategy, detailed in Table \ref{tab:decomposition shapes}, specifically excludes the decomposition of attention layers. This decision was based on observations that decompressing these layers resulted in a more significant performance drop compared to other layers. However, by applying a higher compression ratio to the embedding layer, we managed to achieve a parameter count comparable to DistilGPT-2 and KnGPT-2.
\item \textbf{Knowledge distillation (KD) scheme}: As outlined in Section \ref{knowledge distillation}, we adopt a knowledge distillation strategy for training our compressed model.
\end{itemize}

TQCompressedGPT-2 effectively reduces the parameter count of the original GPT-2$_{Small}$ from 124 million to 81 million. Our comparative analysis includes DistillGPT-2 and setups similar to KnGPT-2, which do not utilize permutation matrices in the decomposition of weight matrices. We maintain the practice of compressing only odd layers of the GPT-2 architecture, a common approach in prior works \cite{kngpt2}. The specific decomposition shapes vary across different setups, with further details provided in the results table. All models were trained using the AdamW optimizer.

One of the key aspects of our approach is the recovery of performance post-compression. By utilizing just a small fraction of the OpenWebText dataset for pre-training, we achieve performance levels on par with KnGPT2 and DistilGPT2. It is important to note that our attempts to replicate the results of the KnGPT2 paper were unsuccessful. Thus, our baseline for comparison is a GPT-2 model compressed using the Kronecker Decomposition method.

\begin{table}[h!]
\centering
\caption{Matrix shapes for GPT-$2_{small}$, and compressed version of it - TQCompressedGPT-2}
\label{tab:decomposition shapes}
\renewcommand{\arraystretch}{2}
  \begin{tabular}{|c|c|c|c|c|}
    \hline 
    Model & Embedding layer & \makecell{Attention layer\\ ($Q, K, V, O$)}& FFN layer & \makecell{Number of \\ parameters} \\
    \hline 
    GPT-2 small & $50527 \times 768$ & $768 \times 768$ & $768 \times 3072$ & 124 mln.\\
    \hline 
    TQCompressedGPT-2 & \makecell{$A: 50527 \times 192$\\ $B: 1 \times 4$} & $768 \times 768$ & \makecell{$A: 768 \times 1536$\\ $B: 1 \times 2$} & 81 mln.\\
    \hline 
  \end{tabular}
\end{table}

\subsection{Results}

Our evaluation of language modeling capabilities is centered around measuring perplexity on the Wikitext-2, Wikitext-103, and Lambada testing subsets. This assessment also includes models without added permutation matrices to compare the effectiveness of our approach.

\begin{table}[h!]
\centering
\caption{Benchmarking results on Wikitext-2, Wikitext-103, and Lambada datasets (lower scores indicate better performance). \\ \textbf{*} Denotes the percentage of the OpenWebText dataset used to train the model.}
\label{tab:experimet results}
\renewcommand{\arraystretch}{1.2}
\begin{tabular}{|c|c|c|c|c|c|}
\hline 
Model & Wikitext-103 & Wikitext-2 & Lambada & Dataset\textbf{*} & \makecell{Number of\\ parameters}\\
\hline
GPT-$2_{small}$ & 29.16 & 24.65 & 45.27 & 100\% & 124 mln. \\
\hline
TQCompressedGPT-2 & \textbf{40.28} & \textbf{32.25} & \textbf{64.72} & 3.1\% & 81 mln. \\

KnGPT-2 & 40.97 & 32.81 & 67.62 & 3.1\% & 81 mln. \\

DistilGPT-2 & 44.53 & 36.47 & 75.99 & 100\% & 82 mln.\\
\hline 
\end{tabular}
\end{table}

%% file: conclusion.tex
\section{Conclusion}
This paper presented a novel approach to compressing neural networks, particularly the GPT-2 model, by introducing a permutation-based enhancement to Kronecker decomposition. Our method stands out for its ability to maintain the structural integrity and performance of the original model while significantly reducing its size. This is achieved through a series of innovative steps: optimal permutations of neuron connections, Kronecker decomposition, and knowledge distillation.

Our TQCompressedGPT-2 model, a compressed version of GPT-2, demonstrates the efficacy of our method. We are comparing the performance of TQCompressedGPT-2 to the original GPT-2$_{small}$, DistillGPT-2 and KnGTP-2. Despite our using only a tiny fraction of the original dataset for KD, we achieve comparable performance levels to  GPT-2$_{small}$. Note that both DistillGPT-2 and KnGTP-2 used far larger portions of the original dataset than we, and yet we have maintained comparable performance. This not only underscores the potential of our approach in practical applications but also opens new avenues for further research. The compressed model, with its fewer parameters, becomes a viable option for deployment in resource-constrained environments.

Looking forward, the implications of this research are far-reaching. Our method can potentially be applied to a wide range of neural network architectures beyond GPT-2, paving the way for more efficient AI models in various domains. The combination of permutation and tensor decompositions presents a new paradigm in neural network compression, balancing the trade-off between model size and performance.

Finally, our research raises intriguing questions about the future of neural network architecture design. The effectiveness of permutations in improving the suitability of neural networks for compression suggests that future architectures could be designed with such optimizations in mind from the outset. As artificial intelligence continues to evolve, techniques like ours will be critical in ensuring that advanced models are accessible and practical for a broader range of applications.

%% file: references.tex
%\section{References}

%% file: appendix.tex
\section*{Appendix}
\subsection*{Appendix 1. Theorem proof.}
\makeatletter\def\@currentlabel{(Appendix 1.)}\makeatother
\label{theorem}
We demonstrate the equivalence of minimizing (3) and solving the assignment problem with matrix (4) as follows:

For any matrix A its Euclidean norm can be presented in such way:

$\|A\|^2 = tr(A^T\cdot A)=tr(A\cdot A^T)$. With respect to that:

\begin{align*}
    \|PW^{(1)} - W^{(2)}\|^2 &= tr[(PW^{(1)} - W^{(2)})(W^{(1)T}P^T - W^{(2)T})] \\
    &= tr(PW^{(1)}W^{(1)T}P^T - PW^{(1)}W^{(2)T} - W^{(2)}W^{(1)T}P^T + W^{(2)}W^{(2)T}) \\
    &= tr(PW^{(1)}W^{(1)T}P^T) - 2tr(PW^{(1)}W^{(2)T}) + tr(W^{(2)}W^{(2)T}) \\
    &= \|W^{(1)}\|^2 - 2tr(PW^{(1)}W^{(2)T}) + \|W^{(2)}\|
\end{align*}

$\|W^{(1)}\|^2 + \|W^{(2)}\|^2=const$. It is equivalent to minimize the function that differs by a constant value, so

$$(\|PW^{(1)} - W^{(2)}\|^2 \rightarrow \min) \Leftrightarrow (tr(P\cdot K) \rightarrow \min)$$

where $K:=-2W^{(1)}W^{(2)T}$.

Here, the initial problem (3) is equivalent to minimization of $tr(P\cdot K)$, where $P$ is a permutation matrix. This is by definition the assignment problem with the matrix $K$.

This explains the values stored in matrix $K$:

\begin{align*}
    (W^{(1)}W^{(2)T})_{ij} &= w^{(1)}_i \cdot w^{(2)T}_j = \sum_k W^{(1)}_{ik}\cdot W^{(2)}_{jk} \\
    K_{ij} &= -2(W^{(1)}W^{(2)T})_{ij} = -2\sum_k W^{(1)}_{ik}\cdot W^{(2)}_{jk}
\end{align*}

The multiplication of $i$-th row of $W^{(1)}$ and $j$-th row of $W^{(2)}$ results in the matrix $K$. Let $D$ be a matrix defined as:

\begin{align*}
    D_{ij} &= \| w^{(1)}_i - w^{(2)}_j\|^2 \\
    &= tr[(w^{(1)}_i - w^{(1)}_j)(w^{(1)T}_i - w^{(2)T}_j)] \\
    &= tr(w^{(1)}_iw^{(1)T}_i) - 2tr(w^{(1)}_iw^{(2)T}_j) + tr(w^{(2)}_jw^{(2)T}_j) \\
    &= \|w^{(1)}_i\|^2 - 2\sum_{k} W^{(1)}_{ik}W^{(2)}_{jk} + \| w^{(2)}_j\|^2.
\end{align*}

As shown, $K_{ij} = - 2\sum_{k} W^{(1)}_{ik}W^{(2)}_{jk} \Rightarrow D_{ij} = \|w^{(1)}_i\|^2 + K_{ij} + \| w^{(2)}_j\|^2$.

$$tr(P\cdot D) =\displaystyle \sum_{i=1}^n D_{i, p(i)} = \sum_{i=1}^n (\|w^{(1)}_i\|^2 + K_{i, p(i)} + \| w^{(2)}_{p(i)}\|^2) = $$ $$ = \sum_{i=1}^n \|w^{(1)}_i\|^2 + tr(P\cdot K) + \sum_{j=1}^n \| w^{(2)}_j\|^2\displaystyle,$$

where $p(i)$ is an index permutation corresponding to permutation matrix $P$.

Again, $\displaystyle\sum_{i=1}^n \|w^{(1)}_i\|^2 + \sum_{j=1}^n \| w^{(2)}_j\|^2 = const$, so minimization of $tr(P\cdot D)$ is equivalent to $tr(P\cdot K)$. Therefore, minimizing (3) is equivalent to the assignment problem with matrix $D$.

%% file: main.bbl
\begin{thebibliography}{15}

\bibitem{kngpt2}
Edalati, A., Tahaei, M., Rashid, A., Nia, V. P., Clark, J. J., \& Rezagholizadeh, M. (2021). Kronecker decomposition for gpt compression. arXiv preprint arXiv:2110.08152.

\bibitem{kroneckerbert}
Tahaei, M. S., Charlaix, E., Nia, V. P., Ghodsi, A., \& Rezagholizadeh, M. (2021). Kroneckerbert: Learning Kronecker decomposition for pre-trained language models via Knowledge Distillation. arXiv preprint arXiv:2109.06243.

\bibitem{glue}
Wang, A., Singh, A., Michael, J., Hill, F., Levy, O., \& Bowman, S. R. (2018). GLUE: A multi-task benchmark and analysis platform for natural language understanding. arXiv preprint arXiv:1804.07461.

\bibitem{tensor_decomposition_in_deep_learning}
Bacciu, D., \& Mandic, D. P. (2020). Tensor decompositions in deep learning. arXiv preprint arXiv:2002.11835.

\bibitem{distilling_the_knowledge_in_neural_network}
Hinton, G., Vinyals, O., \& Dean, J. (2015). Distilling the knowledge in a neural network. arXiv preprint arXiv:1503.02531.

\bibitem{Compression_of_fully-connected_layer_in_neural network by Kronecker product}
Wu, J. N. (2016, February). Compression of fully-connected layer in neural network by Kronecker product. In 2016 Eighth International Conference on Advanced Computational Intelligence (ICACI) (pp. 173-179). IEEE.

\bibitem{hungarian}
Bougleux, S., Gaüzere, B., \& Brun, L. (2017). A {H}ungarian algorithm for error-correcting graph matching. In Graph-Based Representations in Pattern Recognition: 11th IAPR-TC-15 International Workshop, GbRPR 2017, Anacapri, Italy, May 16–18, 2017, Proceedings 11 (pp. 118-127). Springer International Publishing.

\bibitem{svd}
Golub, G. H., \& Reinsch, C. (1971). Singular value decomposition and least squares solutions. In Handbook for Automatic Computation: Volume II: Linear Algebra (pp. 134-151). Berlin, Heidelberg: Springer Berlin Heidelberg.

\bibitem{openwebtext}
Gokaslan, A., \& Cohen, V. (2019) OpenWebText corpus. \url{http://Skylion007.github.io/OpenWebTextCorpus}.

\bibitem{novikov_tensorizing_neural_networks}
Novikov, A., Podoprikhin, D., Osokin, A., \& Vetrov, D. P. (2015). Tensorizing neural networks. Advances in neural information processing systems, 28.

\bibitem{oseledets_tensorizing_embedding_layers}
Hrinchuk, O., Khrulkov, V., Mirvakhabova, L., Orlova, E., \& Oseledets, I. (2019). Tensorized embedding layers for efficient model compression. arXiv preprint arXiv:1901.10787.

\bibitem{distilbert}
Sanh, V., Debut, L., Chaumond, J., \& Wolf, T. (2019). DistilBERT, a distilled version of BERT: smaller, faster, cheaper and lighter. arXiv preprint arXiv:1910.01108.

\bibitem{compression_for_iot devices}
Thakker, U., Beu, J., Gope, D., Zhou, C., Fedorov, I., Dasika, G., \& Mattina, M. (2019). Compressing RNNs for IOT devices by 15-38x using Kronecker products. arXiv preprint arXiv:1906.02876.

\bibitem{kronecker_product_with_applications}
Atteya, M. J. (2016). Kronecker product with applications. MJ J. Algebra Appl., 1, 1-4.

\bibitem{gpt2}
Radford, A., Wu, J., Child, R., Luan, D., Amodei, D., \& Sutskever, I. (2019). Language models are unsupervised multitask learners. OpenAI blog, 1(8), 9.

\bibitem{tensorring}
Zhao, Q., Zhou, G., Xie, S., Zhang, L., \& Cichocki, A. (2016). Tensor Ring decomposition. arXiv preprint arXiv:1606.05535.

\bibitem{tensor_train}
Oseledets, I. V. (2011). Tensor-Train decomposition. SIAM Journal on Scientific Computing, 33(5), 2295-2317.

\bibitem{smoothquant}
Xiao, G., Lin, J., Seznec, M., Wu, H., Demouth, J., \& Han, S. (2023, July). Smoothquant: Accurate and efficient post-training quantization for large language models. In International Conference on Machine Learning (pp. 38087-38099). PMLR.

\bibitem{int8}
Dettmers, T., Lewis, M., Belkada, Y., \& Zettlemoyer, L. (2022). Llm. int8 (): 8-bit matrix multiplication for transformers at scale. arXiv preprint arXiv:2208.07339.

\bibitem{post training pruning}
Kwon, W., Kim, S., Mahoney, M. W., Hassoun, J., Keutzer, K., \& Gholami, A. (2022). A fast post-training pruning framework for transformers. Advances in Neural Information Processing Systems, 35, 24101-24116.

\bibitem{qu2021_tt_hardware}
Qu, Z., Deng, L., Wang, B., Chen, H., Lin, J., Liang, L., Li, G., Zhang, Z. \& Xie, Y. (2021). Hardware-enabled efficient data processing with Tensor-Train decomposition. IEEE Transactions on Computer-Aided Design of Integrated Circuits and Systems, 41(2), 372-385.

\bibitem{lykov2022tensor}
Lykov, D., Schutski, R., Galda, A., Vinokur, V., \& Alexeev, Y. (2022, September). Tensor network quantum simulator with step-dependent parallelization. In 2022 IEEE International Conference on Quantum Computing and Engineering (QCE) (pp. 582-593). IEEE.

\bibitem{kuhn1955hungarian}
Kuhn, H. W. (1955). The Hungarian method for the assignment problem. Naval research logistics quarterly, 2(1‐2), 83-97.

\bibitem{lambada}
Paperno, D., Kruszewski, G., Lazaridou, A., Pham, Q. N., Bernardi, R., Pezzelle, S., Baroni, M., Boleda, G. \& Fernández, R. (2016). The LAMBADA dataset: Word prediction requiring a broad discourse context. arXiv preprint arXiv:1606.06031.

\bibitem{wikitext}
Merity, S., Xiong, C., Bradbury, J., \& Socher, R. (2016). Pointer sentinel mixture models. arXiv preprint arXiv:1609.07843.

\end{thebibliography}
